\documentclass[runningheads]{llncs}

\usepackage{etoolbox}

\AfterEndEnvironment{figure}{\noindent\ignorespaces}
\AfterEndEnvironment{table}{\noindent\ignorespaces}
\AfterEndEnvironment{definition}{\noindent\ignorespaces}
\AfterEndEnvironment{align}{\noindent\ignorespaces}
\AfterEndEnvironment{algorithm}{\noindent\ignorespaces}

\DeclareMathAlphabet{\altmathcal}{OMS}{cmsy}{m}{n}
\usepackage{mathptmx}
\usepackage{amsmath}
\usepackage{amssymb}

\usepackage{graphicx}
\usepackage{tabularx}
\usepackage{xcolor}
\usepackage{multirow}
\usepackage{todonotes}
\usepackage{subcaption}
\usepackage{tikz}
\usepackage{booktabs}

\usepackage[hidelinks]{hyperref}

\makeatletter
\newcommand{\printfnsymbol}[1]{%
  \textsuperscript{\@fnsymbol{#1}}%
}
\makeatother

\newcommand\copyrightnotice{%
    \begin{tikzpicture}[remember picture,overlay]
    \node[anchor=south,yshift=10pt] at (current page.south) {\fbox{\parbox{\dimexpr\textwidth-\fboxsep-\fboxrule\relax}{
    \footnotesize \textcopyright\ Springer, 2020. The final authenticated version is available online at:
\href{https://doi.org/10.1007/978-3-030-76352-7_23}{https://doi.org/10.1007/978-3-030-76352-7\_23}
    }}};
    \end{tikzpicture}%
}

\begin{document}

\title{TELESTO: A Graph Neural Network Model for Anomaly Classification in Cloud Services}

\author{Dominik Scheinert\thanks{equal contribution} \and
Alexander Acker\printfnsymbol{1}}

\titlerunning{TELESTO: A Graph Neural Network Model for Anomaly Classification}
\authorrunning{D. Scheinert \and A. Acker}

\institute{
Distributed and Operating Systems Group, TU Berlin, Berlin, Germany\\
Email: \email{\{firstname.lastname\}@tu-berlin.de}
}

\maketitle   

\begin{abstract}
Deployment, operation and maintenance of large IT systems becomes increasingly complex and puts human experts under extreme stress when problems occur.
Therefore, utilization of machine learning (ML) and artificial intelligence (AI) is applied on IT system operation and maintenance - summarized in the term AIOps.
One specific direction aims at the recognition of re-occurring anomaly types to enable remediation automation.
However, due to IT system specific properties, especially their frequent changes (e.g. software updates, reconfiguration or hardware modernization), recognition of reoccurring anomaly types is challenging.
Current methods mainly assume a static dimensionality of provided data.
We propose a method that is invariant to dimensionality changes of given data.
Resource metric data such as CPU utilization, allocated memory and others are modelled as multivariate time series.
The extraction of temporal and spatial features together with the subsequent anomaly classification is realized by utilizing TELESTO, our novel graph convolutional neural network (GCNN) architecture.
The experimental evaluation is conducted in a real-world cloud testbed deployment that is hosting two applications. 
Classification results of injected anomalies on a cassandra database node show that TELESTO outperforms the alternative GCNNs and achieves an overall classification accuracy of 85.1\%.
Classification results for the other nodes show accuracy values between 85\% and 60\%.

\end{abstract}

%
% ------------------------------------------------
\section{Introduction}
\label{sec:introduction}

The rapid evolution of IT systems enables the development of novel applications and services in a variety of fields like medicine, autonomous transportation or manufacturing.  
Requirements of high availability and minimal latency together with general growth in distribution, size and complexity of these systems aggravate their operation and maintenance.
Human experts require additional support to maintain control and ensure compliance with defined service level agreements (SLAs).
Therefore, monitoring systems are employed to collect key performance indicators (KPIs) like network latency and throughput or system resource utilization from relevant IT system components.
\copyrightnotice
They provide detailed information about the overall system state, which can be used to identify imminent SLA violations.

One specific research direction in that area utilizes methods from machine learning (ML) and artificial intelligence (AI) for operation and maintenance of IT systems (AIOps)~\cite{dang2019aiops,gulenkoa2016system}.
It includes methods for anomaly detection to identify problems ideally before SLAs are violated, anomaly localization to determine the origin of an ongoing anomaly, as well as recommendation and auto-remediation methods to execute actions and transfer anomalous system components back to a normal operation state.
Significant research work is done on methods for anomaly detection~\cite{nedelkoski2019anomaly,wetzig2019unsupervised} and root cause analysis~\cite{wu2020microrca}.
However, existing solutions mostly fail to propose a holistic approach for an automated remediation execution.
This is essential to autonomously transfer anomalous system components back into a normal operation state.
The main reason is the focus on unsupervised methods that are usually trained on one class - the normal state.
During the detection phase deviations from the learned normal class are labeled as anomalies.
Although this is important to enable the detection of previously unseen anomalies, it imposes suboptimal implications.
A generic anomaly class that summarizes all types of anomalies either implies actions that are able to remediate all anomaly types or pushes the responsibility for selecting an appropriate remediation to a subsequent instance - usually a human expert. 
Therefore, we propose a method to recognize reoccurring anomalies by training a classification model.
Utilizing system metric data like CPU utilization, allocated memory or disk I/O statistics and model those as multivariate time series, our model is able to identify anomaly type specific patterns and to assign respective anomaly labels to those.
Currently proposed time series classification methods assume a static dimensionality of input data, which is usually not the case for IT systems, which undergo frequent changes due to software updates, hardware modernization, etc.

To enable the automation of anomaly remediation, we propose a novel anomaly type classification solution, which is utilized to detect reoccurring anomaly types. 
To this end, our proposed model architecture TELESTO utilizes a novel graph neural network architecture to exploit multivariate time series modeled as graphs both in the spatial and temporal dimension.
It is invariant to changing dimensionality and outperforms two other commonly used graph neural network methods.

The rest of the paper is structured as followed. 
In \autoref{sec:approach}, we describe the preliminaries for our approach and present TELESTO in detail. 
A consolidation of the conducted evaluation is given in \autoref{sec:evaluation}, encompassing the hyper-parametrization and training setup, the testbed and experiment design as well as the results of the anomaly classification and their discussion. 
An excerpt of related approaches is presented in \autoref{sec:relatedWork} capturing the state of the art of time series classification in the domain of anomalies. 
Lastly, \autoref{sec:conclusion} concludes this paper and gives an outlook for future work.

\section{Anomaly Classification on Time Series Graphs}

\label{sec:approach}
Our proposed model architecture operates on graphs and utilizes graph convolution to exploit both the spatial and temporal dimension of KPIs modelled as multivariate time series.

\subsection{Preliminaries}
\label{sec:prel}

AIOps systems require monitoring data, which is typically retrieved in form of tracing, logging and resource monitoring metrics.
Latter are usually referred to as key performance indicators (KPIs). 
These can be formally expressed as time series, i.e. a temporally ordered sequence of vectors $X = ({X}_t(\cdot) \in \mathbb{R}^d : t=1,2,\ldots, T)$, where $d$ is the dimensionality of each vector and $T$ defines the last time stamp, at which a sample was observed.
For $X^{a}_{b}(\cdot)=(X_a(\cdot), X_{a+1}(\cdot), \ldots, X_b(\cdot))$, we denote indices $a$ and $b$ with $a \leq b$ and $0 \leq a,b \leq T$ as time series boundaries in order to slice a given series $X^{0}_{T}(\cdot)$ and acquire a subseries $X^{a}_{b}(\cdot)$.
Additionally, we use the notion $X(i)$ to refer to a certain dimension $i$, with $1 \leq i \leq d$.

Our proposed method for anomaly classification relies on modelling time series as graphs.
A Graph $G=(V,E)$ with $n$ nodes consists of a set of vertices $V(G)=\{v_1, \ldots, v_n\}$ and a set of edges $E(G)\subseteq \{\{v_i,v_j\}| v_i,v_j \in V(G)\}$. An edge $\{v_i,v_j\}\in E(G)$ is a connection, i.e. an unordered pair, between vertex $i$ and $j$, thus $v_j$ is called a neighbor of $v_i$ written as $v_i \sim v_j$. The adjacency matrix $A$ of a graph $G$ is an $n \times n$ matrix with entries $a_{ij}$ such that $a_{ij}=1$ if a connection $v_i \sim v_j$ exists, otherwise 0. 

\begin{figure}
    \centering
    \includegraphics[trim=0 80 40 190, clip,keepaspectratio,width=.6\textwidth]{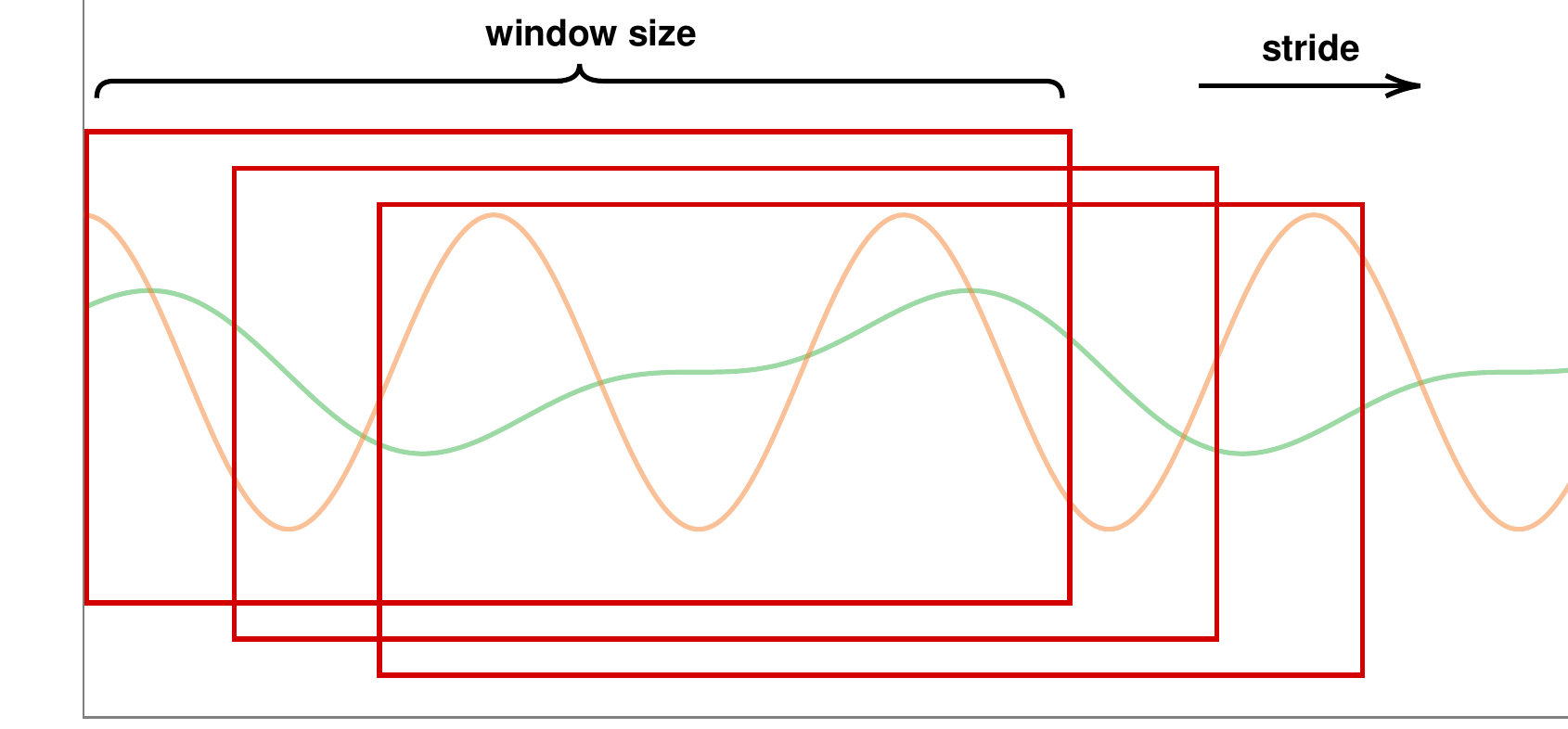}
    \caption{A window is moved along the temporal dimension with a configurable stride while slices of time series data are extracted. Each slice is transformed into a graph.}
    \label{fig:slicing}
\end{figure}

To represent time series as graphs, a sliding window of size $w$ with a configurable stride is moved along the temporal dimension, extracting slices of time series data. 
This is also illustrated in \autoref{fig:slicing}, whereby each red rectangle is transformed into a graph with one node per series. 
Formally, the set of vertices for a graph $G$ at time $t$ is defined as 
\begin{equation}
\label{eq:ts_graph}
\small
     V^{(t)}(G^{(t)}) = \{v_i=\altmathcal{F}(X_{t}^{t-w}(i)) \; | \; i=1, 2, \ldots, d\}.
\end{equation}
Thereby, $\altmathcal{F}$ is a filter, extracting features from time series $X_{t}^{t-w}(i)$.
Edges are used to express the relationship between time series feature vectors and can be either inferred from available data or set manually.
We assume that KPIs where collected during known system states, i.e. either normal or one of a set of known anomaly types $C$.
Therefore, we assign a label $c \in C$ to each graph $G^{(t)}$, defining them as tuples $(G^{(t)}, c)$.

\subsection{The Architecture of TELESTO}
\label{sec:approach_model}
Anomaly classification is done based on multivariate time series modelled as graphs, thus graph classification is required.
Therefore, we employ a class of neural networks which incorporates concepts from graph theory.
Graph convolutional neural networks (GCNNs) aim to generalize the convolution operation to be applied in non Euclidean domains.
We utilize this to model the spatial domain of multivariate time series.
Each node of a graph can have an arbitrary number of neighbors, thus making the method invariant to changing dimensionality. 
The convolution operation is applied on the neighborhood of each graph node. 
GCNN methods can be roughly clustered into spectral and spatial methods. 
Spectral methods are establishing frequency filtering by levering the fourier domain and the graph Laplacian. 
Spacial methods are essentially defining the graph convolution in the vertex domain by leveraging the graph structure and aggregating node information from the neighborhoods in a convolutional fashion. 
A comprehensive survey of existing methods was conducted in~\cite{Zhang2019}.

We propose TELESTO, a novel model architecture for graph classification consisting of multiple spatial methods.
\begin{figure}[!t]
\centering
\includegraphics[keepaspectratio,width=.8\textwidth]{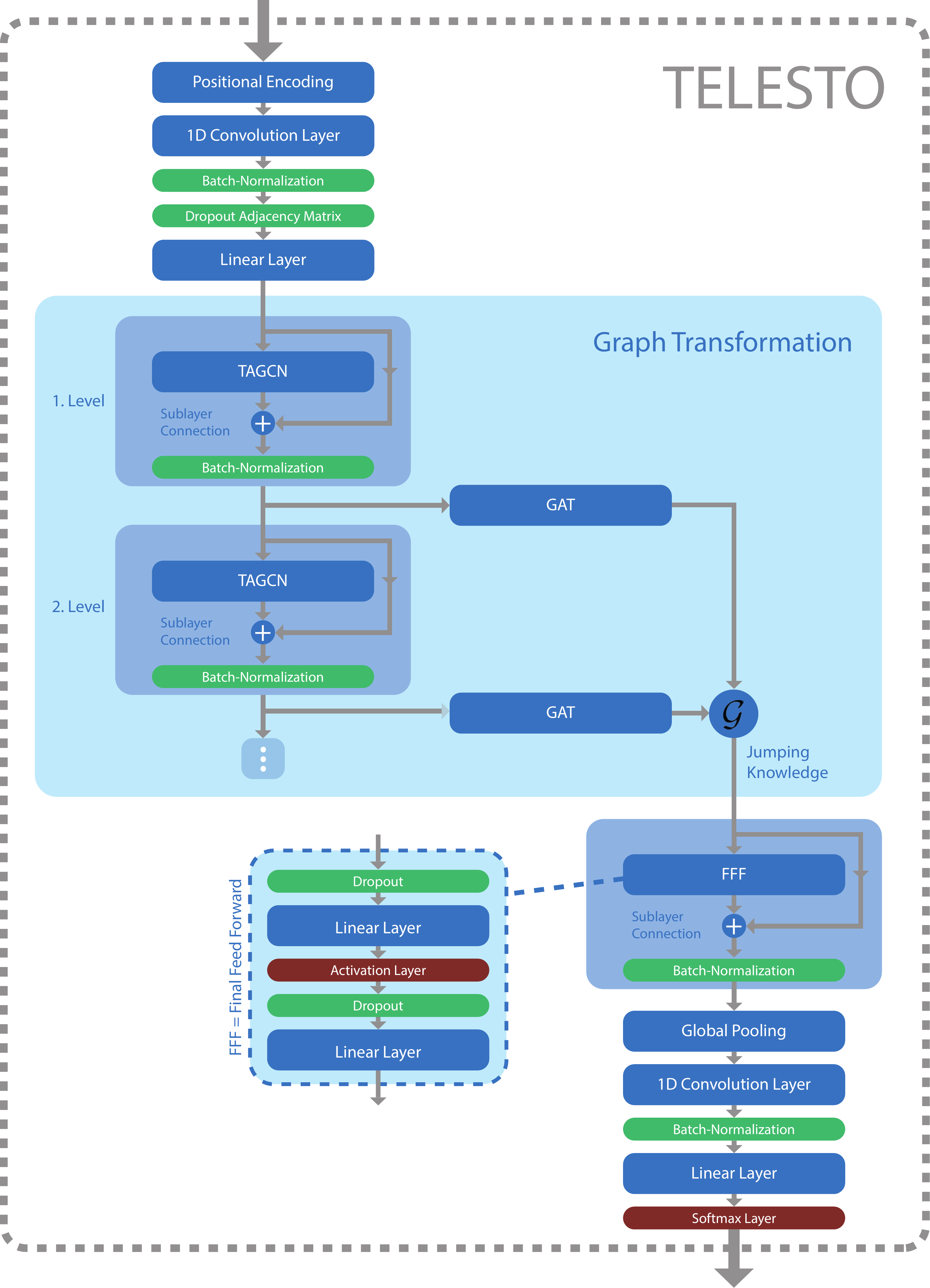}
\caption{The architecture of TELESTO. Green components indicate methods for generalization and red blocks indicate activation functions. An example of a graph transformation module with two blocks is depicted but can be arbitrarily increased.}
\label{fig:model_architecture}
\end{figure}
Our architecture is illustrated in~\autoref{fig:model_architecture}. 
Building upon the definition in~\autoref{eq:ts_graph}, $\altmathcal{F}$ consists of a positional encoding and a subsequent 1D convolution layer.
As argued and evaluated in~\cite{Vaswani2017}, positional encoding allows for the injection of information about the relative or absolute position of values in a sequence.
The convolution layer extracts $N$ features from each time series $X_{t}^{t-w}(i)$.
Thereby a constant filter size of $3$ is used together with a suitable zero-padding to ensure the equivalence of input and output dimensions.
Batch normalization is applied on the features to stabilize the training~\cite{Ioffe2015new}.
Initially, a fully connected graph is used resulting in an all-one adjacency matrix.
A dropout layer randomly sets adjacency matrix entries to zero in order to achieve better generalization~\cite{Srivastava2014}.
The dropout operation is followed by a single linear layer, which allows the model to adjust the dimensionality of the feature vector for the \textit{graph transformation} module, i.e. by mapping from the node feature dimension to the dimensionality of the graph transformation module. 

The graph transformation module is a core component of our approach and is composed of multiple levels / blocks.
On each of its levels, a sublayer (or residual) connection (SLC) is used.
This eases the training of deep neural network architectures~\cite{He2016}. 
For the graph transformation operation, topology adaptive graph convolutional networks (TAGCN)~\cite{Du2017new} and graph attention networks (GAT)~\cite{velivckovic2018graph} are used.
Both aim to generate meaningful node feature representations for the task of anomaly classification.
TAGCNs realize graph convolution on the vertex domain by applying a set of fixed-size learnable filters which are adaptive to the topology of the respective target graph.
The TAGCN result is further used as input for a GAT layer which enables nodes to attend to generate features dependant on the vertex neighborhood features. 
Since we employ GAT layers with multi-head attention, the choice of the number of heads has a direct impact on the output dimensionality of the graph transformation module as the outputs of all parallel attention mechanisms are concatenated.
At each GCNN level, the TAGCN layer exploits the spatial structure of its input graph and respectively updates the node features.
Those are forwarded to the GAT layer as well as to the next TAGCN layer. 
TELESTO allows this combination of TAGCN and GAT layers to be arbitrarily stacked.
Finally, the node feature matrices collected from the respective GAT layers across all levels are aggregated using Jumping Knowledge (JK)~\cite{Xu2018new}. 
JK combines the GAT layer outputs by a configurable aggregator function $\altmathcal{G}$. 
We choose the long short-term memory (LSTM) aggregator, due to its flexibility to learn weight coefficients and thus a weighted combination of the given matrices.
This allows the model to prioritize certain level outputs. 

The graph transformation module is followed by a final feed forward (FFF) block and a sublayer connection.
The FFF block consists of  two dropout layers and two linear layers in alternating order with a single activation layer in between.
Next, graph embeddings are computed by utilizing the global pooling method \textit{Global Attention}~\cite{Li2016new}.
It employs a neural network to learn attention coefficients based on node features, which are then used to aggregate nodes graph-wise, resulting in an embedding for each graph. 
A final 1D convolution is applied on the graph embedding and the result is batch-normalized afterwards. 
The convolution layer is configured to use ten filters, a filter size of nine and suitable zero-padding to ensure the equivalence of input and output dimensions. 
The resulting ten feature maps are averaged. 
Assuming a classification problem with $C$ classes, an additional linear layer needs to be added to the bottom of the architecture to transfer the graph embedding size to an output vector of size $C$. 
A subsequent softmax layer calculates a probability distribution over all classes.
\section{Evaluation}
\label{sec:evaluation}
We investigate a case study based on anomalous services deployed in an infrastructure as a service (IaaS) cloud environment to evaluate our model. Experiments are conducted to assess its goodness by classification of synthetically injected anomalies. 

\subsection{Testbed and Experiment Design} 
\label{sec:evalation_datasets_testbed}
To evaluate the capabilities of our model, we deploy a cloud infrastructure and use a IaaS policy to run two applications.
OpenStack 11.0.0 Stein\footnote{https://www.openstack.org/software/stein/} and Ceph 12.2.5 luminous\footnote{https://docs.ceph.com/docs/master/releases/luminous/} are installed on a commodity cluster of 21 nodes, each possessing an Intel Xeon X3450 CPU (4 cores @ 2.66GHz), 16GB RAM, 3x 1TB HDD and 2x 1GBit Ethernet connection. 
Twelve nodes are used as hypervisors, eight as storage nodes and one as a network and controller node. 
Furthermore, three hypervisor host groups (1/1/10 split) are created to separate the application load generation from the application deployment itself.
A virtual IMS\footnote{https://www.projectclearwater.org} and a content streaming service (CS) are used as example applications hosted within the cloud.
Varying load is generated against both simulating user access.
All VMs operate on Ubuntu 16.04.3 LTS under Linux kernel version 4.4.0-128-generic. 
All deployment scripts related to both services are available at github\footnote{https://github.com/IncrementalRemediation/testbed\_deployment}.

The proposed anomaly classification method requires labeled data to be evaluated.
Therefore, different anomaly types are synthetically injected into the VMs and hypervisors at runtime. 
An injector agent was deployed on each hypervisor and application host VM.
All injected anomaly types are listed in \autoref{table:dataset_anomalies}. 
A group-based injection policy was used throughout the experiments, means that all VMs of one service group (e.g. bono, hypervisor, etc.) are regarded as one entity. 
Note that all ten hypervisors are put into one component group. 
An injection into any component counts as an injection for the whole group. 
During the experiment, an initial period of six hours without anomaly injections is defined. 
Next, each anomaly is injected five times into each group for four to five minutes. 
After that, one minute of grace time is waited until the next injection is performed, i.e. there are no overlapping injections. 
Start and stop times are logged and used as the ground truth. 
Agents are deployed on the hypervisors for monitoring that sample KPIs such as CPU utilization or network I/O statistics at a frequency of 2 Hz. 
\begin{table}
\centering
\caption{List of injected anomalies and their abbreviations together with their respective description and parametrization.}
\small
\begin{tabularx}{.8\textwidth}{
| >{\raggedright\arraybackslash\hsize=.55\hsize}X |
>{\raggedright\arraybackslash\hsize=.45\hsize}X |}
\hline
\textbf{Anomaly} & \textbf{Description} \\ [0.5ex] 
\hline
\begin{tabular}[c]{@{}l@{}}CPU overutilization (CPU)\end{tabular}         & Utilize 90\% of available CPU. \\ \hline
\begin{tabular}[c]{@{}l@{}}Abnormal disk utilization (ADU)\end{tabular}  & \begin{tabular}[c]{@{}l@{}}Constant disk read and write \\ operations. \end{tabular} \\ \hline
Memory leak (MEL)                                                            & \begin{tabular}[c]{@{}l@{}}Incremental allocation of x MB \\ main memory every y seconds.\\ x=1, y=3 for vIMS and CS VMs\\ x=2, y=3 for Hypervisors\end{tabular} \\ \hline
\begin{tabular}[c]{@{}l@{}}Abnormal memory allocation (AMA)                                                           \end{tabular} & \begin{tabular}[c]{@{}l@{}}Allocate x MB of memory.\\ x=450 for vIMS VMs\\ x=900 for CS VMs\\ x=2000 for Hypervisors\end{tabular} \\ \hline
Network overload (NOL)                                                                                                           & Start to download large files.\\ \hline
\end{tabularx}
\label{table:dataset_anomalies}
\end{table}

\subsection{Hyper-Parametrization \& Training Setup}
\label{sec:evaluation_setup}
Monitored KPIs of each node are modelled as multivariate time series $X$.
Each sample of every single serie is $X_t(i)$ is preprocessed by rescaling to the range $(0, 1)$ with min-max normalization.
The limits of most KPIs are well-known (e.g. number of CPU cores and their frequency etc.). 
For some KPIs like latencies between network endpoints, context switches or cache misses, it is challenging to set upper or lower boundaries. 
Those are determined within the training dataset and used throughout testing. 

\begin{table}[ht]
\centering
\caption{Hyper-Parametrization \& Training Setup}
\small
    \begin{tabular}[t]{c|lp{0.6\linewidth}}
        \toprule
        \textbf{Aspect}&\textbf{Configuration}\\
        \midrule
        Hardware & GeForce RTX 2080 Ti GPU\\
        Implementation& PyTorch,\ PyTorch Geometric~\cite{Fey2019new}\\
        Graph Construction& window size = $20$, stride = $1$\\
        Optimizer & Adam~\cite{Kingma2014new} with learning rate = $10^{-3}$, $\beta_1=0.9,\ \beta_2=0.999$\\
        Regularization& weight decay = $10^{-5}$, dropout probability = 50\%\\
        Loss & Cross Entropy\\
        Diverse & Epochs = 15, Batch size = 128, Xavier~\cite{Glorot2010} weight initialization\\
        \bottomrule
    \end{tabular}
\label{tbl:eval_specs}
\end{table}

Most specifications related to the training of TELESTO are listed in \autoref{tbl:eval_specs}. We employ leave one group out (LOGO) cross-validation for data splitting, i.e. the five injections of each anomaly and each service component are split as 3/1/1 as a training/validation/test split. 
For TELESTO itself, we choose a graph node feature dimensionality of 64 and set the number of graph transformation levels to 5. 
For the TAGCN layers, we choose $k=3$ fixed-size learnable filters as recommended in~\cite{Du2017new}. 
For the GAT layers, we choose $K=8$ parallel attention mechanisms to produce rich node features with multi-head attention.
Lastly, the JK LSTM-aggregator is equipped with seven layers in order to learn a reasonable node weighting based on node features. 
We choose ELU~\cite{Clevert2015new} as activation function for the FFF block. 
The final softmax calculates a distribution over anomaly classes, whereof the highest is used as the prediction target.

\subsection{Anomaly Classification}
\label{sec:evaluation_classification}
In this section, the proposed model architecture will be evaluated on the data described in~\autoref{sec:evalation_datasets_testbed}.
TELESTO and its default configuration is compared against a GCN architecture~\cite{Kipf2017new} and a GIN architecture~\cite{Xu2019new}.
GCN is a reasonable choice as it is a common benchmark, whereas GIN is selected due to it achieving state-of-the-art results both for node classification and graph classification on several benchmark data sets~\cite{Xu2019new}. 
All models are trained on each service node individually.
Moreover, each experiment is run 10 times and the results are averaged in order to cancel out the effects of unfavorable weight initialization.
Both the GCN architecture and the GIN architecture utilize two of their respective layers. For graph classification, the node features are added across the node dimension for each graph, followed by two linear layers with dropout in between and a final softmax layer.  
Specific to the GCN architecture is the hidden layer size of 32 and the row-normalization of input feature vectors. 
For the GIN architecture, the hidden layer size is set to 64 while for each GIN layer, the input is batch-normalized, the initial value of $\epsilon$ is set to 0 and a multilayer perceptron (MLP) is internally used for mapping the node features from the input dimension to the hidden dimension. If not specified otherwise, we use the values from the setup summarized in \autoref{tbl:eval_specs}.
\begin{table}[t!]
\centering
\caption{The results of anomaly classification on the cassandra data set.}
\small
\begin{tabular}{c||c|ccccc||c}
\hline
\textbf{Model} & \textbf{Metric} & \textbf{Split 1} & \textbf{Split 2} & \textbf{Split 3} & \textbf{Split 4} & \textbf{Split 5} & $\varnothing$\\
\hline
\hline
\multirow{4}{*}{GCN} & Accuracy & 0.389 & 0.463 & 0.360 & 0.398 & 0.385 & 0.399\\
& Recall & 0.389 & 0.462 & 0.358 & 0.396 & 0.385 & 0.398\\
& Precision & 0.272 & 0.428 & 0.256 & 0.222 & 0.288 & 0.293\\
& F1-Score & 0.310 & 0.436 & 0.295 & 0.269 & 0.320 & 0.326\\
\hline
\multirow{4}{*}{GIN} & Accuracy & 0.447 & 0.455 & 0.562 & 0.452 & 0.400 & 0.463\\
& Recall & 0.446 & 0.455 & 0.562 & 0.452 & 0.399 & 0.463\\
& Precision & 0.408 & 0.422 & 0.569 & 0.363 & 0.289 & 0.410\\
& F1-Score & 0.421 & 0.436 & 0.564 & 0.402 & 0.335 & 0.432\\
\hline
\multirow{4}{*}{TELESTO} & Accuracy & 0.796 & 0.804 & 0.894 & 0.822 & 0.939 & 0.851\\
& Recall & 0.796 & 0.803 & 0.894 & 0.822 & 0.939 & 0.851\\
& Precision & 0.825 & 0.732 & 0.920 & 0.870 & 0.956 & 0.861\\
& F1-Score & 0.810 & 0.764 & 0.906 & 0.844 & 0.948 & 0.854\\
\hline                
\end{tabular}
\label{table:model_comparison}
\end{table}

A detailed breakdown of the results is given in \autoref{table:model_comparison} with a focus on the cassandra service node. It can be seen that the proposed model outperforms both comparative models. 
Therefore we conclude that the proposed model is suitable for anomaly classification based on multivariate time series data. 
Note that for TELESTO, the average F1-score of 0.854 is close to its reported average accuracy. 
In general, this is an indication for a good model as the balanced recall and precision are not strongly different from the accuracy.
In contrast to that, both comparative models appear to encounter difficulties during training.
The GCN architecture achieves an average F1-score of 0.326 whereas the GIN architecture performs comparably better with an average F1-score of 0.432. 
It can be observed that the reported F1-scores differ from the achieved average accuracy of these architectures. 
\autoref{table:model_comparison} also shows a high variability between splits. 
For instance, TELESTO achieves an average F1-score of 0.906 on split 3 but an average F1-score of 0.764 on split 2.
Moreover, an investigation of the corresponding confusion matrices shows that confusion exists between CPU anomalies and MEL anomalies as well as AMA anomalies and MEL anomalies. 
Similar observations regarding diverse F1-scores can be made for all models between multiple splits. 
One possible explanation might be the high variability in simulated user load during our experiments. 
This resulted in a broad range of system states, from almost idle to almost over utilized. 
With the overall available training data being limited, such high load variability might lead to significant differences in classification performances between splits. 
For TELESTO, we observed that although the validation accuracy proportionally increases with the training accuracy, the losses on both data sets structurally diverge after a few epochs already, leading to an overfitting of the model which is intensified by the models complexity.

For completeness, we report the results of anomaly classification with TELESTO on all other service nodes in \autoref{table:model_results} while reporting only accuracy scores to omit redundancy.
\begin{table}[t!]
\centering
\caption{The results of anomaly classification with TELESTO on all service nodes. The table shows the achieved accuracy scores for each split and in average across all splits.}
\small
\begin{tabular}{l|ccccc||c}
\hline
\textbf{Data Set} & \textbf{Split 1} & \textbf{Split 2} & \textbf{Split 3} & \textbf{Split 4} & \textbf{Split 5} & $\varnothing$\\
\hline
\hline
cassandra & 0.796 & 0.804 & 0.894 & 0.822 & 0.939 & 0.851\\
bono & 0.865 & 0.853 & 0.825 & 0.566 & 0.729 & 0.768 \\
sprout & 0.818 & 0.797 & 0.660 & 0.567 & 0.597 & 0.688\\ 
backend & 0.655 & 0.792 & 0.714 & 0.912 & 0.731 & 0.761\\
chronos & 0.806 & 0.741 & 0.984 & 0.898 & 0.620 & 0.810\\
homer & 0.214 & 0.529 & 0.725 & 0.846 & 0.677 & 0.598\\
astaire & 0.630 & 0.716 & 0.931 & 0.702 & 0.744 & 0.744\\
load-balancer & 0.801 & 0.880 & 0.989 & 0.767 & 0.680 & 0.823\\
homestead & 0.539 & 0.599 & 0.855 & 0.730 & 0.744 & 0.694\\
\hline                
\end{tabular}
\label{table:model_results}
\end{table}
It can be seen that the performance of TELESTO varies across different service nodes and splits. 
While an average accuracy of 0.851 can be achieved on cassandra, the average accuracy on homer is 0.598. 
The reported scores also strongly vary between splits on homer, between 0.214 average accuracy on split 1 and 0.846 on split 4. While not listed in  \autoref{table:model_results}, the comparative models exhibit high variance in accuracy over splits and remain on average significantly inferior to TELESTO.

\subsection{Limitations}
\label{sec:evaluation_limitations}
The main problem encountered during our extensive experimentation is the contradiction between the expected amount of labeled anomaly data and required data to train a reliable model.
Labeling anomaly data by human experts is costly, anomalies occur sparsely and IT environments undergo constant changes so once labeled data deprecates over time.
Therefore, the generalization ability  of our model represented by the prediction scores reveal a significant variance in dependence of a specific split.
We plan to investigate methods and heuristics to generate additional training data from few anomaly examples and thus, synthetically increase the available training data size.

Another aspect is the lack of expression regarding the temporal dependence between consecutive graphs.
Although sequential information of time series within a graph are encoded via positional encoding, consecutive graphs constructed via a moving window over the time series are regarded as independent.
Possibilities to encode temporal information from preceding graphs for the classification of subsequent graphs is subject to future work.

\section{Related Work}
\label{sec:relatedWork}
Concrete methods of identifying reoccurring anomalies within IT systems is sparsely covered on public research.
We formulate it as a time series classification problem. 
Therefore, we analyse related work in both areas, IT system anomaly classification and time series classification in general.

Bodik et al.~\cite{bodik2010fingerprinting} referring to the classification of different data center crises as data center fingerprinting.
Thereby, system resource metrics from all data center components are aggregated via quantile discretization and feature selection methods are applied to choose relevant metrics for distinguishing different data center crisis types.
The reported results were achieved by an aggregation of system resource metrics collected over 30 minutes.
Kajó and Nováczki~\cite{kajo2016genetic} provide a comparison of different machine learning algorithms together with a genetic algorithm approach. 
Given monitoring data from a System Architecture Evolution (SAE) core network they select an optimized metric subset that is used as input for different classification algorithms. 
The focus lies on metric selection and classification models are trained on 850 anomaly observations. 
Having sparse occurrences of anomaly situations the expectation of many training data is a major limit.
Cheng et al.~\cite{cheng2018multi} applies a multi-scale long short-term memory model to classify four different anomaly types based on update messages of the border gateway protocol (BGP). 
The approach expects the classification models to be trained with several hours of anomaly data.
However, anomaly situations usually do not persist for an extended period of time in production systems.

A variety of time series classification exists. We focus on most recent published approaches.
% #####################
% ### InceptionTime ###
% #####################
InceptionTime \cite{Fawaz2020new} is an ensemble of deep CNNs for time series classification. 
Each CNN consist of multiple inception modules, whereas every module utilizes bottleneck layers for regularization and applies both a sliding max-pooling operation and multiple sliding filters of different lengths for feature extraction. 
In InceptionTime, multiple architecturally equivalent networks with different initial weight values are utilized and their prediction outputs are evenly weighted to obtain a final prediction result.
% ################
% ### LSTM-FCN ###
% ################
Another approach which incorporates the advantages of LSTM networks is named LSTM-FCN \cite{Karim2017}. 
It consists of two parallel processing streams. 
In the first stream, the temporal structure of the input data is exploited by an LSTM module. 
The second stream leverages alternating convolution layers, batch normalization layers and activation layers and a final global average pooling (GAP) layer.
In the end, the concatenation of both stream outputs is used for classification.
% #############
% ### T-GCN ###
% #############
Another method named T-GCN is presented in \cite{Zhao2019}. 
The model combines graph convolutional networks together with GRU and thus aims at capturing the spatial and temporal dependencies simultaneously. 
The T-GCN model can be seen as an improvement of LSTM-FCN, since a convolution is applied that is not bound to the euclidean domain.
After that, the convolution result is processed by a recurrent neural network (RNN). 
Although a recent publication, T-GCN utilizes a graph convolution method that is outperformed by other GCNN models.

 \section{Conclusion}
 \label{sec:conclusion}
In this paper we presented TELESTO, a novel time series classification model to identify reoccurring anomalies in services deployed in a IaaS cloud environment.
Therefore, we model KPIs of hypervisors and virtual machines that are hosting applications as multivariate time series.
A method to transform multivariate time series into graphs is presented.
The proposed model is based on GCNNs and thus, invariant to changes of the input dimensionality.
We apply convolution on both the spatial and temporal dimension to extract a set of features that are used for classifying anomalies via graph classification.
To evaluate the method, a cloud system together with two applications hosted within an IaaS service model were deployed.
Synthetic injections of anomalies provided the required ground truth for evaluation.
TELESTO was able to outperform two state of the art GCNNs, revealed promising results for anomaly classification and thus, is able to detect reoccurring anomalies in services deployed in cloud environments.

For future work we want to examine ways to encode temporal information from preceding graphs for the classification of subsequent graphs.
Further, different time series augmentation methods can be tested to synthetically increase the amount of data.
% ################################
% ################################
\bibliographystyle{splncs04}
\small\bibliography{all_publications}

\end{document}